\documentclass{article}

\usepackage{arxiv}
\usepackage{amsmath}
\usepackage[utf8]{inputenc} 
\usepackage[T1]{fontenc}    
\usepackage{hyperref}       
\usepackage{url}            
\usepackage{booktabs}       
\usepackage{amsfonts}       
\usepackage{nicefrac}       
\usepackage{microtype}      
\usepackage{cleveref}       
\usepackage{lipsum}         
\usepackage{graphicx}
\usepackage{natbib}
\usepackage{doi}
\usepackage{url,hyperref,lineno,microtype,subcaption}
\usepackage[onehalfspacing]{setspace}
\usepackage{multirow}
\usepackage{tabularx}
\usepackage{stmaryrd}
\usepackage{hyperref}

\title{Fusarium head blight detection, spikelet estimation, and severity assessment in wheat using 3D convolutional neural networks}

\author{ Oumaima ~Hamila
		\\
	Department of Applied Computer Science\\
	The University of Winnipeg\\
	Winnipeg, MB, Canada \\
	\texttt{hamila-o@webmail.uwinnipeg.ca} \\
	\And
	Christopher J. ~Henry\\
	Department of Applied Computer Science\\
	The University of Winnipeg\\
	Winnipeg, MB, Canada\\
	\texttt{ch.henry@uwinnipeg.ca} \\
	\And
	Oscar I. ~Molina\\
	Morden Research and Development Centre\\
	Agriculture and Agri-Food Canada\\
	Morden, MB, Canada\\
	\texttt{oscar.molina@agr.gc.ca} \\
	\And
	Christopher P. ~Bidinosti\\
	Department of Physics\\
	The University of Winnipeg\\
	Winnipeg, MB, Canada\\
	\texttt{c.bidinosti@uwinnipeg.ca} \\
	\And
	Maria Antonia ~Henriquez\\
	Morden Research and Development Centre\\
	Agriculture and Agri-Food Canada\\
	Morden, MB, Canada\\
	\texttt{mariaantonia.henriquez@agr.gc.ca} \\
}

\renewcommand{\headeright}{}

\renewcommand{\shorttitle}{}

\hypersetup{
pdftitle={Fusarium head blight detection, spikelet estimation, and severity assessment in wheat using 3D convolutional neural networks},
pdfauthor={Oumaima ~Hamila},
pdfkeywords={Fusarium head blight, wheat,  severity index, convolutional neural networks, 3D, detection, estimation, multispectral point cloud},
}

\begin{document}
\maketitle

\begin{abstract}
	Fusarium head blight (FHB) is one of the most significant diseases affecting wheat and other small grain cereals worldwide. The development of resistant varieties requires the laborious task of field and greenhouse phenotyping. The applications considered in this work are the automated detection of FHB disease symptoms expressed on a wheat plant, the automated estimation of the total number of spikelets and the total number of infected spikelets on a wheat head, and the automated assessment of the FHB severity in infected wheat. The data used to generate the results are 3-dimensional (3D) multispectral point clouds (PC), which are 3D collections of points – each associated with a red, green, blue (RGB), and near-infrared (NIR) measurement. Over $300$ wheat plant images were collected using a multispectral 3D scanner, and the labelled UW-MRDC 3D wheat dataset was created. The data was used to develop novel and efficient 3D convolutional neural network (CNN) models for FHB detection, which achieved $100\%$ accuracy. The influence of the multispectral information on performance was evaluated, and our results showed the dominance of the RGB channels over both the NIR and the NIR plus RGB channels combined. Furthermore, novel and efficient 3D CNNs were created to estimate the total number of spikelets and the total number of infected spikelets on a wheat head, and our best models achieved mean absolute errors (MAE) of $1.13$ and $1.56$, respectively. Moreover, 3D CNN models for FHB severity estimation were created, and our best model achieved $8.6$ MAE. A linear regression analysis between the visual FHB severity assessment and the FHB severity predicted by our 3D CNN was performed, and the results showed a significant correlation between the two variables with a $0.0001$ P-value and $0.94$ R-squared.
\end{abstract}


\section{Introduction}
Fusarium head blight (FHB) is a devastating fungal disease caused by a variety of species within the \emph{Fusarium} genus. Although it mainly affects wheat, it can also affect other cereals like barley and oats \citep{agr_paper13}. The development of FHB is favourably influenced by wet, moist, and warm weather conditions. The fungus infects the spikes \citep{spikelet} during the flowering stage, causing a deficiency in the plant's development and premature grain shivering and bleaching, which results in significant yield losses in quality and quantity. Moreover, trichothecene mycotoxins, such as deoxynivalenol (DON), may be triggered and accumulated in the infected kernels which causes acute toxicity to both humans and animals \citep{FHB_toxic}. Additionally, in Canada, the severity and frequency of the disease have been increasing every year \citep{agr_paper14}, and the annual reported losses range from $\$50$ to $\$300$ million since $1990$. Therefore, FHB is considered a serious food safety and economic issue that calls for efficient and safe solutions for disease identification, control, and prevention. 

To help reduce the impact of FHB or mycotoxin contamination, many practices and management strategies are being adapted by farmers and researchers. Examples include using crop rotation, tillage, variety selection, and fungicide use. Among these practices, the development of wheat cultivars with resistance to FHB is considered a high priority worldwide and a major bottleneck for wheat breeding programs \citep{ijms21124497}. In order to develop resistant cultivars, multiple wheat varieties are seeded, grown, inoculated with fungus, and then tested for their level of resistance, which can be characterized by the percentage of spikelets\footnote{A head (also known as a spike) consists of a number of spikelets, and a spikelet consists of florets that could develop into $1–3$ grains.} on the infected wheat head with visually detectable disease. This FHB severity percentage is determined by dividing the total number of infected spikelets by the total number of spikelets on the same wheat spike and then multiplying the sum by $100$, as illustrated in Figure \ref{methodology} (H). However, this percentage is typically calculated by a visual observation carried out by human agents, which results in a subjective assessment that is prone to inaccuracy. Moreover, a daily assessment of the FHB severity for thousands of wheat plants in indoor growth chambers or wheat fields is a very demanding and time-consuming task that involves many agents and requires high levels of expertise, concentration, and accuracy. These issues necessitate the creation of automated tools that can replace the arduous manual tasks of visually identifying FHB symptoms, counting the number of infected spikelets per wheat head, and determining the FHB severity of diseased wheat.

The technological advances in multispectral and hyperspectral imaging, remote sensing, and 3-dimensional (3D) imaging that occurred in agriculture during recent years \citep{agr_paper6,agr_paper7} have led to the development of advanced acquisition systems such as drones and scanners that are used to create 2D and 3D image datasets of plants and crops \citep{agr_paper8}. Meanwhile, machine learning (ML) has drastically evolved due to the advance of computing power, the availability of large labelled datasets, and new algorithms with many more parameters than were previously computationally possible \citep{machineLearning}. For these reasons, many advanced applications in digital agriculture were created, such as yield monitoring \citep{agr_paper9}, plant disease detection \citep{agr_paper10}, and monitoring FHB wheat using hyperspectral imagery and unmanned aerial vehicles \citep{UAV_hyper}. However, despite the fact that 3D data is growing in popularity in view of the advantages it provides \citep{3dImagingSystems}, such as employing an extra spatial dimension to represent the depth of an image and thereby increase the amount of information \citep{10.5555/971144}, there are currently very few publications in the literature that use 3D data in digital agriculture applications in contrast to 2D data. The main reasons for the underuse of 3D data are its scarcity and the high computational costs associated with processing it.

One of the most commonly used types of 3D data are point clouds (PCs), which are collections of points scattered in a 3D space and represent the 3D shapes of objects. These data points are often each associated with colour information, such as red, green, and blue (RGB) measurements. PCs give detailed representations of objects in a 3D space by providing a more realistic description of the objects’ edges, surfaces, and textures than their 2D counterparts, which, for example, distort the information about the depth of a real object when projecting it onto a flat surface. \citep{10.5555/971144}. Despite the numerous advantages of PCs, there are not many PC datasets of wheat \citep{agriculture11060563,agriculture11050450}. As a result, in this study, we created a novel labelled PC dataset \citep{SP3/QJWBEM_2023}. The dataset was created through a collaboration of the TerraByte research group\footnote{\href{https://acs.uwinnipeg.ca/terrabyte/}{terrabyte.acs.uwinnipeg.ca}} at the University of Winnipeg (UW) and Agriculture and Agri-Food Canada's Morden Research and Development Centre (MRDC). For this reason, we have labelled the dataset reported here as the UW-MRDC 3D Wheat Dataset. The dataset consists of Dataset I and Dataset II, which both represent water-controlled (WC) (\emph{i.e.} healthy) and FHB-infected PCs of wheat, and each were used for different applications.
The main distinction between datasets I and II is the protocol followed for the plant inoculation. All data within the UW-MRDC 3D wheat dataset was acquired using a multispectral 3D scanner, which produces a PC combined with multispectral information, such that each point within the PC is associated with RGB and NIR intensities detected at that point.

Several published works explored the detection of FHB in wheat using image processing techniques or deep learning models with hyperspectral or multispectral images of wheat kernels \citep{agr_paper1}, field wheat \citep{10.3389/fpls.2022.1004427}, wheat spikes or heads \citep{ALMOUJAHED2022107456,agr_paper2}, or wheat seeds \citep{agriculture12111801}. In this work, Dataset I was used to develop 3D convolutional neural networks (CNNs), which are deep learning models that excel at mapping complex input data to specific class labels, for the detection of FHB symptoms in wheat that would serve as an automated assessment tool for scientists who conduct research on wheat in labs or indoor growth chambers. 



A few studies that attempted to automate the counting of wheat spikes or kernels were developed. Examples include an Android app that counts wheat grains based on image segmentation \citep{10.3389/fpls.2022.821717} and a phenotyping system that uses image processing and deep learning to count the number of spikelets in a lab setting \citep{10.3389/fpls.2022.872555}. Other studies attempted to directly estimate the FHB severity without determining the total number of spikelets or the total number of infected spikelets. Examples include an approach that used pre-trained models and transfer-learning techniques to estimate FHB severity from 2D images of wheat \citep{agronomy12081876} and a method based on extracting texture and colour features from hyperspectral images of wheat heads for use in training classical machine learning models (such as support vector machines) to predict the severity of the FHB \citep{s20102887}. Despite the findings of these studies, there are currently no studies in the literature that automate the counting of the number of spikelets, the number of infected spikelets, or the FHB severity in wheat using PCs or 3D images. 
Therefore, in this study, we developed 3D CNN models to automatically estimate the FHB severity of diseased wheat using Dataset II. Moreover, we developed two 3D CNN models, one of which automatically estimates the total number of spikelets and the other estimates the total number of infected spikelets using Dataset I for wheat heads and Dataset II, respectively. Since the FHB severity is the ratio of the total number of infected spikelets in a wheat head to the total number of spikelets on the same wheat head, these two models were created to be used as an alternate technique to estimate the FHB severity of infected wheat by dividing the two predictions.

The main contributions of this work are (i) a novel labelled dataset called the UW-MRDC 3D Wheat dataset that represents multispectral PCs of wheat plants consisting of both healthy and FHB-diseased samples. The dataset consists of Dataset I, which contains two collections of PCs, the first of which represents wheat spikes and the second of which represents wheat heads, and Dataset II, which contains PCs that represent wheat spikes, (ii) a real-time CUDA-based preprocessing model for the conversion of multispectral PCs into multispectral 3D images, (iii) an accurate, reliable, and real-time 3D CNN model for FHB detection in wheat from multispectral 3D images, (iv) the empirical determination of the most important spectral information for FHB detection with CNNs, (v) an efficient 3D CNN model for estimating the total number of spikelets in a wheat head, (vi) an efficient 3D CNN model for estimating the total number of infected spikelets in a wheat head, (vii) an efficient 3D CNN model for estimating FHB severity in in wheat, and (viii) a linear regression analysis between the visual FHB-disease assessment and the assessment predicted by the 3D CNN.

\section{Materials and methods}

\subsection{Methodology overview}

\begin{figure}[h]
\begin{center}
\includegraphics[width=180mm]{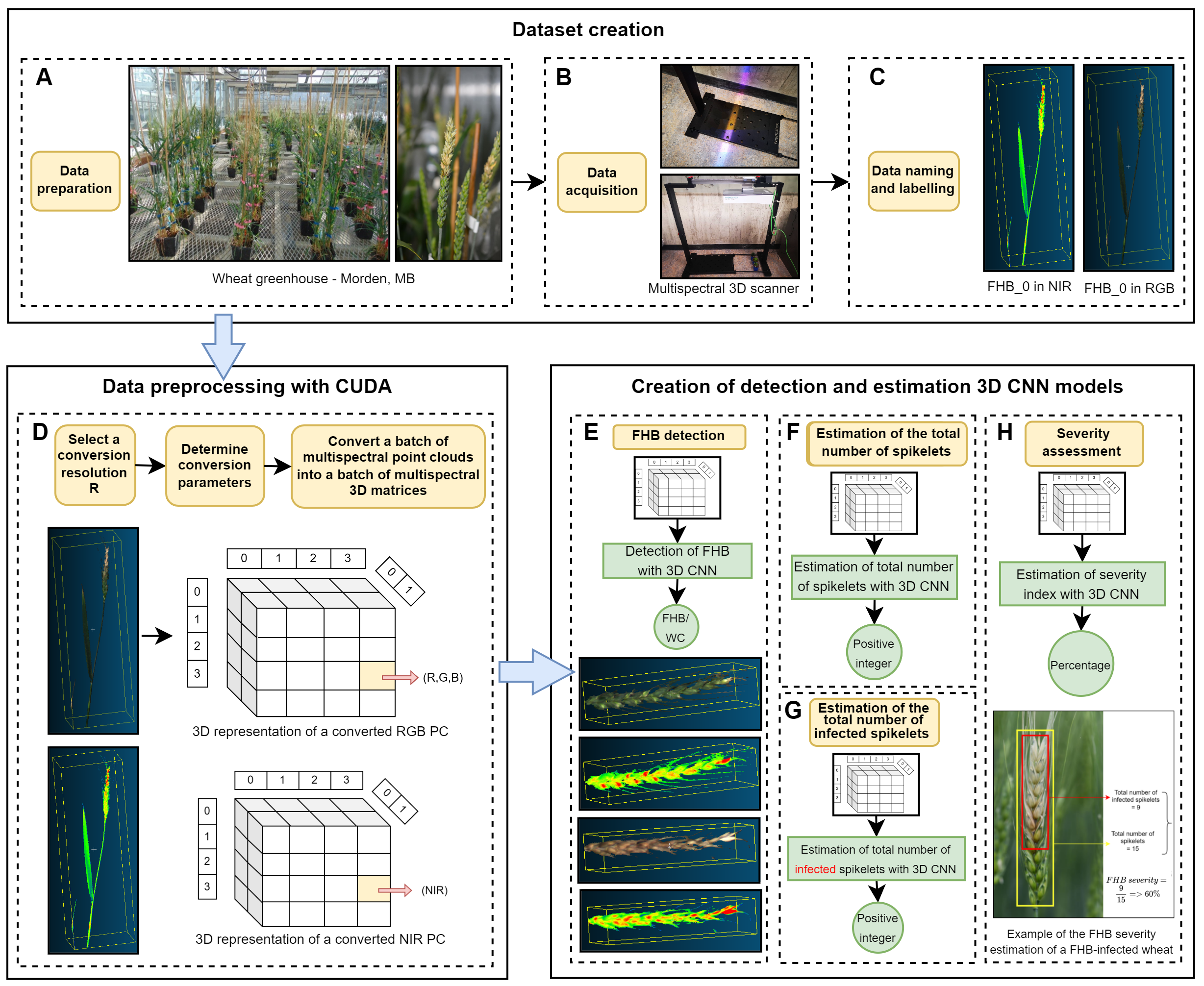}
\end{center}
\caption{Methodology overview of this study which consists of three major components: (i) dataset creation (A, B, and C), (ii) data preprocessing with CUDA (D), and (iii) creation of detection and estimation 3D CNN models (E, F, G, and H). These methods were created to achieve FHB detection (E), spikelet estimation (F and G), and severity assessment (H) in wheat using 3D convolutional neural network models and multispectral point cloud data.}\label{methodology}
\end{figure}

The overall procedure that was designed and developed to conduct this study is shown in Figure \ref{methodology}. It consists of three major systems: dataset creation, data preprocessing, and model creation for detection and estimation in wheat using 3D CNNs. The dataset creation process consists of the three steps depicted in Figure \ref{methodology}: data preparation (A); data acquisition (B); and data naming and labelling (C). UW-MRDC 3D wheat dataset, that consists of datasets I and II, is the dataset created to conduct this study \citep{SP3/QJWBEM_2023}.
Following dataset creation is data preprocessing with CUDA (D), during which data samples that are multispectral PCs were transformed into multispectral 3D images, whose representation is compatible with CNNs. Finally, following data preprocessing is model creation, in which 3D CNNs were developed and trained to automate the tasks of FHB detection (E); total number of spikelets estimation (F); total number of infected spikelets estimation (G); and severity assessment in wheat (H). For the development of the (E) and (F) applications, Dataset I was used, whereas Dataset II was used for (G) and (H). 

\subsection{UW-MRDC 3D wheat dataset creation}

\subsubsection{Dataset overview}

\begin{figure}[h]
\begin{center}
\includegraphics[width=180mm]{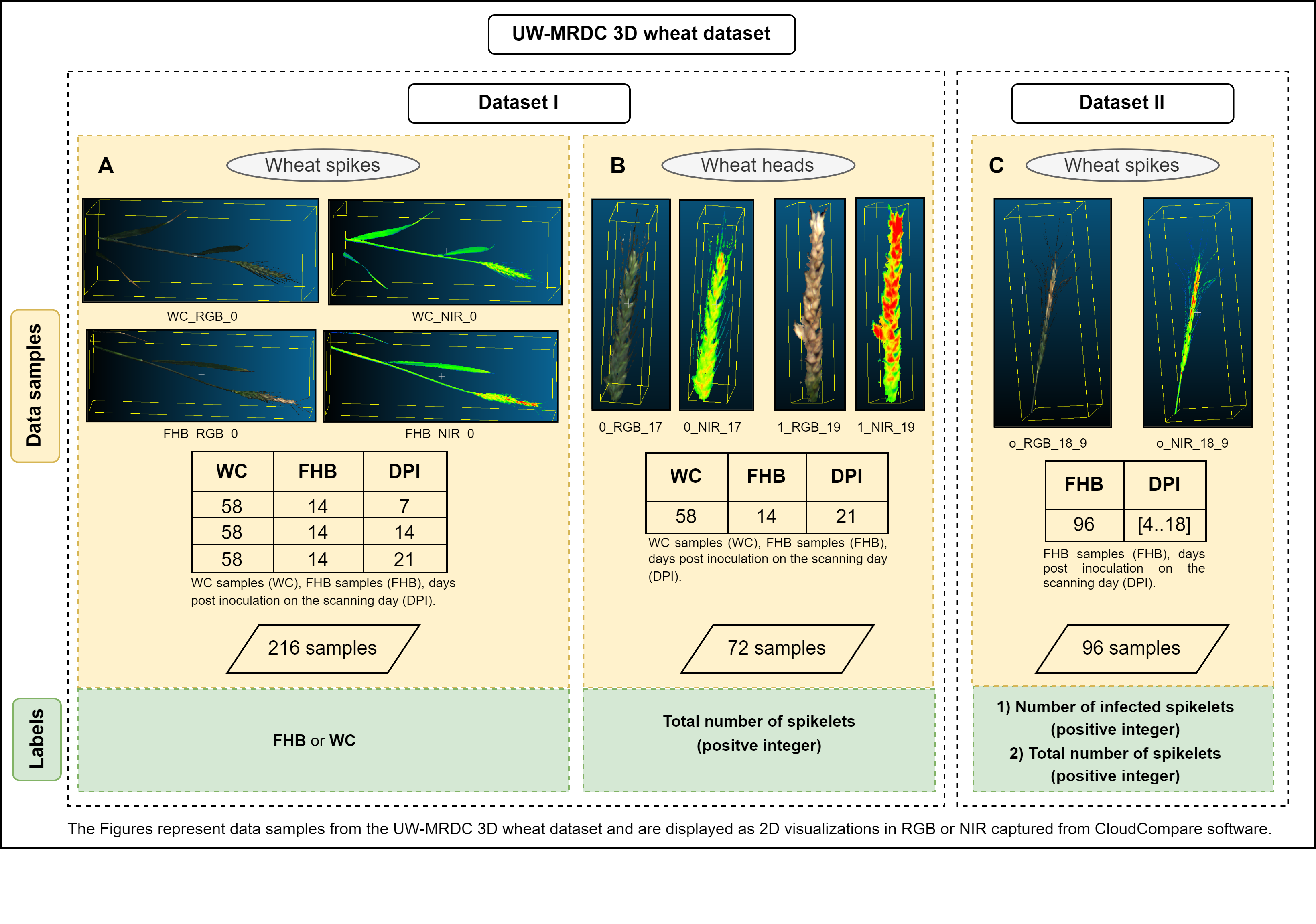}
\end{center}
\caption{Overview of the content and specifications of the UW-MRDC 3D wheat dataset consisting of Dataset I and II. Dataset I consists of two collections of PC data: wheat spikes (A) and wheat heads (B), and Dataset II consists of one collection of PC data: wheat spikes (C).}\label{datasets}
\end{figure}

Figure \ref{datasets} illustrates the content of the UW-MRDC 3D wheat dataset, which consists of two collections of data. The first collection was called Dataset I, while the second collection was called Dataset II. The main difference between the two collections is the methods used during the data preparation phase, which is described in detail in Section \ref{data_preparation}. Dataset I (A) represents wheat spikes. It was created to achieve the task of FHB detection in wheat. The dataset was acquired by scanning $72$ wheat plants, of which $14$ were inoculated and $58$ were kept WC, at three different growth stages. All $72$ plants were captured at $7$, $14$, and $21$ days post-inoculation (DPI), representing the growth stages of $7$ days after Zadoks $65$ (seven days after anthesis), Zadoks $73$ (early milk), and Zadoks $83$ (early dough), respectively.
Plants were scanned on different dates to capture the development of disease symptoms over time. Early FHB symptoms were recorded when at least one wheat spikelet turned yellow or pinkish and became distinguishable from the other green spikelets, and, as time went by, the disease kept developing and more spikelets got infected in a wheat head. The final Dataset I for wheat spikes consists of $216$ labelled PCs, where each PC is labelled either FHB or WC. Dataset I (B) represents wheat heads. It was created to achieve the task of total number of spikelets estimation; therefore, the scans were focused only on the wheat spikes (also called wheat heads). The data was acquired by cutting the wheat stems of the $72$ wheat plants  at $21$ DPI and then scanning the remaining heads. The ensemble of the dataset consists of $72$ PCs, each labelled by a positive integer in the range ${\llbracket}7,22{\rrbracket}$. Finally, Dataset II (C) for wheat spikes was created to achieve the tasks of estimating the total number of infected spikelets and the FHB severity. Therefore, only FHB-diseased wheat with visible symptoms was captured at different DPIs ranging from $4$ to $18$ DPI, as shown in Table (C) in Figure \ref{datasets}. The final dataset consists of $96$ PCs, each of which is labelled with two positive integer values, the first of which indicates the total number of spikelets and ranges between ${\llbracket}13,21{\rrbracket}$ and the second of which reflects the number of infected spikelets in a wheat head and ranges between ${\llbracket}2,15{\rrbracket}$.
\subsubsection{Data preparation}
\label{data_preparation}

The plant material used for Dataset I included the Canada Western Red Spring (CWRS) wheat cultivar 5602HR and CDC Teal. The 5602HR cultivar is moderately resistant to FHB, and CDC Teal is susceptible. The plant material used for Dataset II included only the wheat cultivar 5602HR. Planting and inoculation methods are identical to those described in Nilsen et al. \citep{Nilsen}. A 3-acetyldeoxynivalenol producing isolate of \emph{Fusarium graminearum} (\emph{Fg}) (HSW-$15$-$39$), obtained from the Henriquez Spring Wheat (HSW) collection of \emph{Fusarium} isolates, was used for Dataset I. For Dataset II, ten 3-acetyldeoxynivalenol (3-ADON) and ten $15$-acetyldeoxynivalenol ($15$-ADON) producing isolates of \emph{Fg} were used in this study. In summary, seeds were sown in $3.5$" pots with a mixture of $50\%$ Sunshine soilless \#$5$ mix (manufactured by Sun Gro Horticulture) and $50\%$ soil, plus $6$ g of slow-release Osmocote $14$-$14$-$14$ fertilizer (manufactured by the Scotts Company). Plants were grown in controlled-environment cabinets with $16$ hours of light at $22$°C and $8$ h of darkness at $15$°C. The bilateral florets of a spikelet positioned at the fifth spikelet in the upper part of a spike were inoculated at $50\%$ anthesis with a $10$ µL of \emph{Fg} macroconidia suspension ($5$×$10^{4}$ macroconidia/mL), which was performed between the lemma and palea using a micro-pipette. Control plants were treated with sterile water. Inoculated plants were covered with a plastic bag for $48$ hours to promote infection. FHB severity was calculated by counting the number of spikelets showing disease symptoms within each spike at $7$, $14$, and $21$ DPI.

\subsubsection{Data acquisition}
All the wheat plants in this work were scanned using Phenospex's PlantEye F500 multispectral 3D scanner \citep{3D_scanner}. It captures data non-destructively and delivers 3D representations of plants (via PCs) in real time. A wheat plant container is placed beneath the PlantEye, which, once activated, begins moving forward and emitting multispectral light beams onto the plant. The reflections of those beams are acquired to form a 3D representation of the wheat plant that include intensities of the four different color bands (RGB and NIR). Multispectral information and 3D representation are then combined into a single PC. The light wavelength of the PlantEye ranges in nanometers (nm) from $[460,750]$, and the peak wavelengths in nm of the blue, green, red, and NIR channels are $[460,485]$, $[530,540]$, $[620,645]$, and $[720,750]$, respectively. 

\subsection{Data preprocessing with CUDA}
The multispectral PCs generated from the PlantEye 3D scanner are stored in a polygon file format known as PLY, which is a file format designed specifically to save 3D models. A PLY file contains tuples of flat polygons in addition to tuples of colour information. Flat polygons and colour information are described by a tuple of $(x,y,z)$ coordinate values varying between negative and positive floating-points and a tuple of (R,G,B,NIR) intensity values, where each value is stored as an integer varying between $[0,255]$. Moreover, the tuple of point coordinates stored in a PLY file is unordered, such that each point is independent and unrelated to the remaining points within the tuple. The ensemble of points is useful to reconstruct 3D models in space by placing each coordinate in its specific spatial position. However, PLY representation does not support complex operations such as convolutions and matrix manipulations that require points within a data signal to be correlated and organised such that a meaningful change in space or time between points can be defined. As a result, in this work, a C++ programme that converts PLY files into 3D images and that runs on GPUs was developed to overcome the limitations of using CNNs on PLY files. The CUDA \citep{cuda} parallel computing platform and programming model was employed to develop the conversion model. 
\subsubsection{Theory and implementation of point cloud to 3D image conversion}
Our proposed solution for converting PLY files into 3D images was based on linear interpolation, such that every point coordinate in the tuple of points stored in a PLY file was converted through linear interpolation into a new voxel coordinate within the constructed 3D image. This interpolation was necessary because the coordinates stored in a PLY file can be either negative or positive floating points, while the coordinates required by 3D CNNs have to be positive integers for indexing. The conversion operations are repetitive and separable, meaning that they can be applied independently to all the point coordinates in the PLY file, which provided a perfect opportunity to exploit GPU parallelism. Thus, the proposed CUDA-based method applies the same linear operations simultaneously on all the points in a PLY file. The conversion equations defining the linear interpolation along the $x$, $y$, and $z$-axes are
\begin{equation}
\label{eq1}
\begin{split}
x_{matrix} &= \left\lceil a_{x} \ x_{PC}+b_{x} \right\rceil,\\
y_{matrix} &= \left\lceil a_{y} \ y_{PC}+b_{y} \right\rceil,\\
z_{matrix} &= \left\lceil a_{z} \ z_{PC}+b_{z} \right\rceil,\\
\end{split}
\end{equation}
such that $a_{x}$, $a_{y}$, and $a_{z}$ are respectively the function slope corresponding to the $x$, $y$, and $z$-axes, and $b_{x}$, $b_{y}$, and $b_{z}$ are respectively their intercepts. $x_{matrix}$, $y_{matrix}$, and $z_{matrix}$ are the positions of the point along the width, height, and depth of the output 3D image, corresponding respectively to the transformation of $x$, $y$, and $z$ values of a point coordinate in the PLY file, noted respectively as $x_{PC}$, $y_{PC}$, and $z_{PC}$. Moreover, $\lceil x \rceil$ defines the ceiling function of a real number $x$ that is defined as the smallest integer that is not smaller than $x$. The function's slopes and intercepts are calculated as 
\vspace{3mm}
\begin{equation} 
\label{eq2}
\begin{split}
a_{x} & = \frac{\left\lceil R(\max_{1\leq i\leq N}(x_{i})-\min_{1\leq i\leq N}(x_{i}))\right\rceil}{\max_{1\leq i\leq N}(x_{i})-\min_{1\leq i\leq N}(x_{i})}, b_{x} = -a_{x}\min_{1\leq i\leq N}(x_{i}),\\
a_{y} & = \frac{\left\lceil R(\max_{1\leq i\leq N}(y_{i})-\min_{1\leq i\leq N}(y_{i}))\right\rceil}{\max_{1\leq i\leq N}(y_{i})-\min_{1\leq i\leq N}(y_{i})}, b_{y} = -a_{y}\min_{1\leq i\leq N}(y_{i}),\\
a_{z} & = \frac{\left\lceil R(\max_{1\leq i\leq N}(z_{i})-\min_{1\leq i\leq N}(z_{i}))\right\rceil}{\max_{1\leq i\leq N}(z_{i})-\min_{1\leq i\leq N}(z_{i})}, b_{z} = -a_{z}\min_{1\leq i\leq N}(z_{i}),\\
\end{split} 
\end{equation}
such that $N$ is the total number of points in the tuple of point coordinates in the PLY file, $(x_i,y_i,z_i)$ is the coordinate of the $i^{th}$ point in the tuple, and $R$ is the resolution factor that serves to enlarge or reduce the resolution of the output 3D image. Finally, for the linear transformation, only the spatial coordinates $(x,y,z)$ were used to estimate the new voxel coordinates ($x_{matrix},y_{matrix},z_{matrix}$), while their corresponding colour intensities (R,G,B,NIR) were reallocated in the new voxel coordinates within the 3D image. The dimensions of the output 3D image are
\vspace{2mm}
\begin{equation} 
\label{eq3}
\begin{split}
width & = \left\lceil R(\max_{1\leq i\leq N}(x_{i})-\min_{1\leq i\leq N}(x_{i}))\right\rceil,\\
height & = \left\lceil R(\max_{1\leq i\leq N}(y_{i})-\min_{1\leq i\leq N}(y_{i}))\right\rceil,\\ 
depth & = \left\lceil R(\max_{1\leq i\leq N}(z_{i})-\min_{1\leq i\leq N}(z_{i}))\right\rceil,\\ 
\end{split} 
\end{equation}
such that width, height, and depth correspond to the range of values along the $x$, $y$, and $z$-axis, respectively.

\begin{figure}[h]
\begin{center}
\includegraphics[width=180mm]{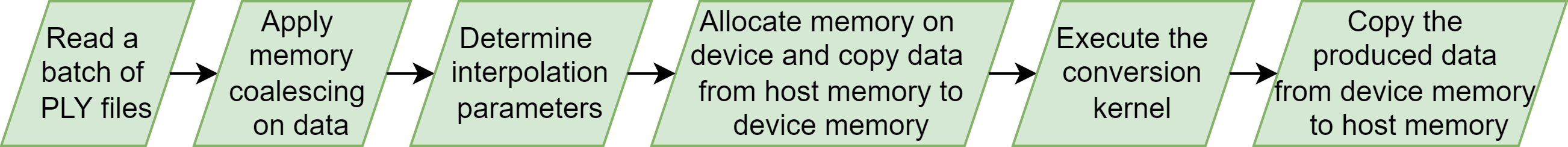}
\caption[Diagram of CUDA implementation steps.]{Diagram of CUDA implementation steps to convert a batch of PLY files into a batch of 3D images.}
\label{cuda}
\end{center}
\end{figure}

To implement Equations \ref{eq1}, \ref{eq2}, and \ref{eq3}, a general C++ API called hapPLY \citep{hapPLY} was used to load PLY files. The API allows the reading and writing of the properties of a PLY file, such as the point coordinates and their corresponding colour intensities, and loads them as two separate tuples of real-values. Figure \ref{cuda} shows the implementation steps followed in the CUDA code to convert a batch of PLY files into a batch of 3D images. The code started by reading the properties of a batch of PLY files. Processing the data in batches allows for further optimization of parallel execution with CUDA, such that the code processes all data points of $n\times$PCs simultaneously, rather than only the data points of a single PC. Next, all elements of the tuples of coordinates and tuples of colours of the $n\times$PCs were rearranged in a manner that ensures memory coalescing (see section \ref{Memory_coalescing} for more details), which enables accessing consecutive memory locations within a single I/O operation. Following that, the maximum and minimum values of the coordinates needed to estimate the parameters of the interpolation functions and the dimensions of the output batch of 3D images were determined and used for calculations. Next, the memory space needed for the data that was used during the kernel execution was allocated on the device memory, and the data was copied from the host memory to the device memory. Then, the conversion kernel, which is the function executed on the GPUs, was launched to convert the batch of PLY files into their corresponding 3D images. Finally, the produced batch of 3D images was copied from the device memory to the host memory.

\subsubsection{Memory coalescing}
\label{Memory_coalescing}
With respect to the CUDA programming model, threads within a thread block are organized into warps, where a warp is a group of $32$ consecutive threads assigned to execute the same set of operations. In practice, threads within a warp access sequential memory locations for read and write operations. This means that memory access operations can be a major bottleneck for GPU applications if the data accessed by sequential threads in a warp is not sequentially ordered in memory. Therefore, the solution is memory coalescing \citep{cuda}, which is a technique used by CUDA where global memory accesses of threads in a warp are grouped together into one operation to minimize global memory bandwidth. In fact, each time a global memory location is accessed, a set of consecutive locations, including the requested location, are also accessed. Thus, in order to reduce the latency caused by data access operations, we made sure that the data used by consecutive threads in a warp is stored in consecutive memory locations. 

\begin{figure}[h]
\begin{center}
\includegraphics[width=85mm]{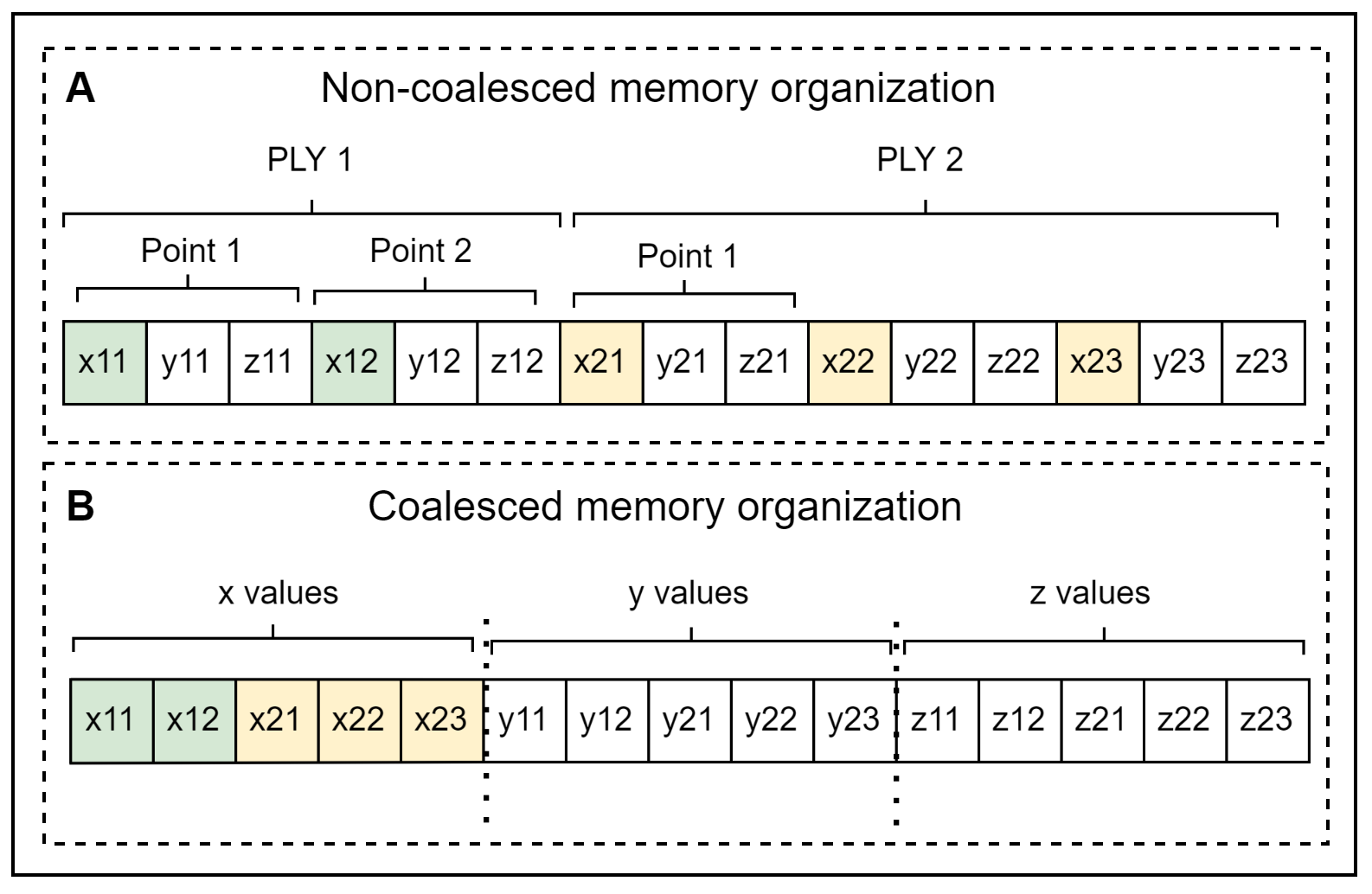}
\caption{Data organisation in a memory array for both coalesced and non-coalesced patterns of data. (A) depicts the raw storage of data in a memory array, and (B) depicts a the storage of the same data in memory in a coalesced manner.}
    \label{coalesced}
\end{center}
\end{figure}

The kernel that performs the data conversion operations was programmed to estimate each value within the tuple of point coordinates separately, which means that $x_{matrix}$, $y_{matrix}$, and $z_{matrix}$ are all estimated independently of one another. Thus, threads within a block were designed such that each block of threads was programmed to load and operate on either the $x_{PC}$, $y_{PC}$, or $z_{PC}$ values to calculate either the $x_{matrix}$, $y_{matrix}$, or $z_{matrix}$ values, respectively. The kernel architecture was designed to take advantage of memory coalescing during data loading, so that one thread within a warp loads all consecutive $x_{PC}$, $y_{PC}$, or $z_{PC}$ values from memory into the cache, allowing the remaining threads to load their corresponding data directly from the cache and execute their operations faster. Figure \ref{coalesced} shows the data organisation in a memory array for both coalesced and non-coalesced patterns. The illustrated examples use only a few points per PC for the purpose of demonstration only. Figure \ref{coalesced}(A) depicts the raw storage of data in a memory array in which point coordinates corresponding to a batch of PLY files are arranged in such a way that points corresponding to the first file are stored first, followed by points corresponding to the second file, and so on, and each point is stored by its $(x,y,z)$ coordinates, where each memory slot contains one coordinate value. The first two point coordinates represent the first PLY file of the batch, while the following points in the array correspond to the second PLY file within the same batch. $(x11,y11,z11)$ represents the first point of the first PLY file within the batch, followed by $(x12,y12,z12)$ which represents the second point of the first PLY file. Once all the points corresponding to the first PC within the batch are stored, the points corresponding to the second PC are added to the same array. In the example, $(x21,y21,z21)$ represents the first point of the second PLY file, and so on. This kind of arrangement is not suitable for an optimised CUDA kernel execution because the memory accesses will be inefficient. 

Thus, point coordinates in memory were rearranged to ensure that threads access coalesced data locations during kernel execution. Figure \ref{coalesced}(B) shows an array where all the $x$ values representing all the points from the PLY files in a batch that were stored consecutively are placed in successive memory slots, followed by all the $y$ values, and finally all the $z$ values. Not only were the tuple of point coordinates rearranged to support data coalescing, but also the tuple of colours. Colour intensities were loaded such that each (R,G,B,NIR) tuple corresponding to the first point of the first PLY file was the first element of the memory array, followed by the second (R,G,B,NIR) tuple corresponding to the first PLY file and so on. Thus, the tuples of colours were rearranged so that all $R$ values representing the points of the first PLY file within a batch were put first in the memory array, followed by all the $R$ values of the second PLY file, and so on. Once the $R$ values were stored, $G$, $B$, and NIR values were then stored consecutively in the memory array according to the same memory coalescing principle.

\subsubsection{Conversion kernel}
\label{conversion_kernel}
The conversion kernel function was implemented to perform point coordinate transformations from their original spatial placement within the PC to their new voxel positions within the 3D dimensions of a 3D image. Each thread was designed to calculate the linear interpolation of a single point, which means that each thread executed the linear interpolation functions defined in Equation \ref{eq1} and related to the $(x, y, z)$ values of a point's coordinates. Firstly, the number of threads allocated on the device memory was determined to be $\frac{1}{3}$ of the coordinates list, and those threads were each programmed to execute three linear interpolations related to their designated $(x, y, z)$ coordinates in order to determine the new voxel coordinates within the output 3D image. Next, each thread loaded the $(R,G,B)$ colour intensities and placed the tuple of colours in their corresponding voxel position within the output 3D image. In fact, the interpolation functions defined in Equation \ref{eq1} convert floating-point coordinates into integer coordinates (with the ceiling operation) that define the voxel positions within the constructed 3D image. Moreover, in some cases, more than one real-valued point coordinate may get converted into the exact same voxel coordinate. In that case, the newer point would override the existing one, resulting in a reduction in the total number of points defined in the 3D image. Furthermore, the size of the constructed 3D image, as defined in Equation \ref{eq3}, ensured that the object defined in the PC was converted into a minimum bounding box, which was the generated 3D image. Moreover, the voxel values that remained empty after reassigning the colour tuples from their positions in the PC to their new voxel positions within the constructed 3D image, were set to zero. The conversion kernel described in this section produced 3-channel 3D images with each voxel value consisting of a tuple of $(R,G,B)$ colour intensities, while NIR intensity values were processed through a second kernel to produce 1-channel 3D images.

\subsection{Model development for the detection of fusarium head blight}
\label{fhb_detection}

\subsubsection{Monitored grid search}
\label{monitored_grid_search}
In our study, 3D CNNs were developed from scratch. A grid search over the number of layers and the number of neurons per layer was conducted. The objective of the grid search was to find the optimal 3D CNN architecture that produces the highest accuracy on the task of FHB detection from 3D images of wheat. The layers employed to build the models were 3D convolution layers, 3D max pooling layers, and densely-connected (or dense) layers, and the search space used to determine the optimal number of layers and neurons was the following:
\begin{itemize}
\item[-] \{$3,4,5,6$\}: Search space of the number of 3D convolution + 3D max pooling layers. The last layer of 3D convolution before the densely-connected layers is not followed by a 3D max pooling layer. 
\item[-] \{$1,2,3,4,5,6$\}: Search space of the number of densely-connected layers.
\item[-] \{$128,64,32,16,8$\}: Search space of the number of neurons per layer. The last densely-connected layer has always one neuron.
\end{itemize}
However, with these sets of variables, the number of possible combinations is $380{,}835{,}000$ networks, which is too large to search exhaustively. Thus, a monitored grid search was employed as an alternative to training only a small number of 3D CNN models. Hence, the monitored grid search worked by randomly generating a batch of $20$ networks at a time, such that a 5-fold cross-validation (CV) \citep{CrossValidation} was performed on each network in the batch, and then the top three networks that achieved the highest average CV accuracy out of the $20$ networks were retrained on the training set and evaluated on the test set. 

\subsubsection{Datasets Characteristics}
\label{data_characteristics}
Three different dataset versions were used in this application, each used to train the $20$ 3D CNN networks. The datasets were obtained by converting Dataset I of wheat spikes into 3D images with a resolution factor $R=1$. However, the datasets differed by their voxel information, which were defined as
\begin{itemize}
\item[1.] The 3D wheat-plant images in RGB (3DWP\_RGB): In this dataset, the voxels of the 3D images contained RGB colour information  ($3$ channels). 
\item[2.] The 3D wheat-plant images in NIR (3DWP\_NIR): In this dataset, the voxels of the 3D images contained NIR colour information ($1$ channel).
\item[3.] The 3D wheat-plant images in RGB${+}$NIR (3DWP\_RGB\_NIR): In this dataset, the voxels of the 3D images contained the RGB${+}$NIR colour information ($4$ channels).  
\end{itemize}

3D images within the datasets had different sizes, such that the width, height, and depth values corresponding to the 3D images dimensions were within ${\llbracket}25,237{\rrbracket}$, ${\llbracket}85,378{\rrbracket}$, and ${\llbracket}14,384{\rrbracket}$, respectively. Since CNNs require input samples of a fixed size, resizing the totality of the 3D images within the datasets to the same size was required. The easiest option was to resize every 3D image to the maximum size that corresponded to $237\times 378\times 384$ voxels (vx). However, this method raised the volume of the data tremendously, such that each resized 3D image contained $34{,}401{,}024$ vx. Training the models on big volume data consumes too much time and resources. Thus, a smaller fixed size was determined such that, any batch of resized 3D images could fit in the GPU memory along with any of the aforementioned model parameters. Since the height dimension of a 3D image represented the real height of a scanned wheat, it was important to preserve it as much as possible when resizing. Therefore, a fixed size was determined by fixing the height to $300$ and by calculating the width and depth via the average aspect ratios of the images in the datasets. As a result, the data samples were all resized to a fixed size equal to $75\times 300\times 95$ vx while maintaining their respective original aspect ratios. This means that a 3D image was resized to the highest possible size that preserved the initial height--width proportion, preserved the height--depth proportion, and that was contained within the $75\times 300\times 95$ vx envelope. The resized images were then zero--padded to $75\times 300\times 95$ vx. 

\subsubsection{5-fold cross validation}
\label{5fcv}
Prior to training the $20$ models, the data samples were divided into $90\%$ training set and $10\%$ test set. Since FHB detection is a binary classification, we ensured that the training set and the test set had the same class distribution with respect to FHB and WC classes. Next, the training samples were further split into $5$ folds that had the same class distribution as the training set to perform the CV, such that each fold consisted of $20\%$ of the training data. Then, a 5-fold CV was applied by training the models on the four training folds and validating them on the validation fold. The top three model architectures that achieved the highest average CV accuracy were retrained on the entire training set and evaluated on the test set, where "average CV accuracy" refers to the average accuracy value achieved by the network trained on each fold of the five CV folds.

\subsubsection{Model Architectures}
\label{modelsArchitectures}

In Table \ref{models_20}, the architectures of each of the $20$ models that were constructed by the monitored grid search are presented. The number of convolutional neurons refers to the number of neurons per 3D convolutional layer and the number of fully-connected neurons refers to the number of neurons per fully-connected layer. Even though the architectures of the models were randomly generated through the monitored grid search, only architectures with a descending order of the number of neurons per both convolution layers and fully connected layers were considered valid candidate models. In other words, given a layer $l$ with a number of neurons equal to $n_{l}$, the number of neurons $n_{l+1}$ in the direct subsequent layer $l+1$ had to be less than or equal to the number of neurons in layer $l$. (\emph{i.e.} $n_{l+1} \le n_{l}$). The choice of the decreasing number of neurons throughout the layers created lighter models with a relatively small and condensed number of parameters. Every 3D convolutional layer was followed by a 3D max pooling layer except for the last 3D convolutional layer. The activation function in all the layers was the rectified linear unit (ReLU) \citep{ReLU}, except for the output layer, where the activation was a sigmoid function. By default, the last fully-connected layer has $1$ neuron since the networks were solving a detection problem. For instance, model $1$ had three 3D convolutional layers such that the number of neurons per layer from layer $1$ to $3$ are equal to $16$, $8$, and $8$, respectively, and had four fully-connected layers such that the number of neurons per layer from layer $1$ to $4$ are equal to $128$, $64$, $8$, and $8$, respectively. The top three 3D CNN models with the highest average CV accuracy on the 3DWP\_RGB dataset were models 8, 10, and 11. While models 8, 9, and 11 and models 3, 5, and 9 were the top three models that achieved the highest average CV accuracy on the 3DWP\_RGB\_NIR dataset and the 3DWP\_NIR dataset, respectively.

\begin{table}[h]
\caption[Architectures of the $20$ 3D CNN models created from the monitored grid search.]{Overall architectures of the $20$ 3D CNN models created from the monitored grid search for the detection of FHB-disease symptoms in wheat. \# of convolutional neuron is the number of neurons per convolutional layer, and \# of fully-connected neurons is the number of neurons per fully-connected layer.}
\label{models_20}
\centering
\begin{tabular}{lcc}
\noalign{\smallskip}\noalign{\smallskip} 
\textbf{Model} & \textbf{\# of convolutional neurons} & \textbf{\# of fully-connected neurons} \\\hline
\noalign{\smallskip}
1 & 16,8,8 & 128,64,8,8 \\
2 & 64,64,64,32,8 & 128,32,8 \\
3 & 32,32,8,8 & 128,64,32 \\
4 & 64,64,16 & 128,128,32 \\
5 & 32,16,16,8 & 32,16 \\
6 & 64,64,64,16 & 16 \\
7 & 64,16,16,16 & 32,64,16 \\
8 & 32,32,32,32,16 & 128,64,32,16 \\
9 & 32,32,32,16,16 & 64,32,16,8 \\
10 & 32,32,32,8,8 & 64,32,16 \\
11 & 32,32,32,32,16 & 128,64 \\
12 & 16,8,8,32,64 & 32 \\
13 & 64,64,8,8,8 & 32,16 \\
14 & 64,32,8 & 128,32,16,8 \\
15 & 64,64,32 & 128,16,8 \\
16 & 64,16,8,8 & 16,8 \\
17 & 32,32,32,16 & 128 \\
18 & 32,32 & 8,8,8 \\
19 & 32,32,16,16,8 & 32 \\
20 & 64,32,32 & 128 \\
\hline
\end{tabular}
\end{table}

\subsubsection{Model Training}
To train each model, a batch size equal to $5$ was used because there was not enough memory on the GPU to store a bigger batch. The root mean square propagation (RMSProp) \citep{RMSProp} optimization algorithm was used to update each network's parameters, with a learning rate equal to $5e^{-4}$. And, binary cross entropy (CE) loss function \citep{BCE} was used as the loss function. Each model was trained for $100$ epochs. To implement the 3D CNN models, we used the Python programming language and its open-source neural network library Keras \citep{keras}. We conducted the experiments on a NVIDIA Tesla P100 GPU server with 12GB of GPU memory.

\subsection{Model development for the estimation of the total number of spikelets}
To calculate the FHB severity, one option is to estimate the ratio components, which are the total number of spikelets and the total number of infected spielets. Thus, it was essential to create accurate and efficient CNN models that produce reliable predictions of these two quantities in order to achieve accurate FHB severity estimations. In this Section, we developed 3D CNN for the estimation of the total number of spikelets, including healthy and diseased ones.

\subsubsection{Monitored grid search and predefined models adaptation }
Two approaches were followed to create regression models for the estimation of the total number of spikelets. In the first approach, 3D CNN networks were created from scratch through a monitored grid search. In the second approach, three well-known CNN architectures were adapted for use with 3D data to solve the regression problem. These three networks were deep residual learning (ResNet) \citep{Resnet} in two variations (ResNet v1 and ResNet v2 \citep{thesis}), and densely connected convolutional networks (DenseNet) \citep{densenet, thesis}.

In the first approach, a monitored grid search over the number of layers and the number of neurons per layer was used to build five 3D CNN models for estimating the total number of spikelets. The monitored grid search used the same search space described in Section \ref{monitored_grid_search}. 

In the second approach, 3D ResNet v1, 3D ResNet v2, and 3D DenseNet models were created by adapting the ResNet v1, ResNet v2, and DenseNet models. 3D ResNet v1 and 3D ResNet v2 were created by transforming every 2D convolutional layer and 2D average pooling layer into a 3D convolutional layer and a 3D average pooling layer, respectively. Moreover, the activation function of all the output layers was changed from a sigmoid function to a ReLU function. Similarly, a 3D DenseNet was created by changing every 2D convolutional layer and 2D average pooling layer in DenseNet into a 3D convolutional layer and a 3D average pooling layer, respectively. Furthermore, the activation function of the output layer was changed from a sigmoid function to a ReLU function. In total, two 3D ResNet v1 networks, three 3D ResNet v2, and two 3D DenseNet were created. 

\subsubsection{Datasets characteristics and labels}
The dataset used for this application was Dataset I of wheat heads. The PCs were converted using a resolution factor of $1.5$ and, only 3D images with RGB colour information were used. Then, all the 3D images were resized into $(161\times 51\times 93)$ vx, which corresponded to the maximum width, height, and depth within the dataset samples. The labels of the dataset samples were integers that vary between $7$ and $22$.

\subsubsection{5-fold Cross Validation}
\label{5cv_2}
Prior to training, the samples were divided into $90\%$ training set and $10\%$ test set. Despite the problem being unrelated to the FHB disease and indifferent to the health of the wheat head, the equality of the class distribution data in all the splits was ensured in terms of FHB and WC samples. Next, the training samples were further divided into $5$ balanced folds to perform 5-fold CV, such that each fold consists of $20\%$ of the
training data. Then, a 5-fold CV was performed on each model, meaning that a 5-fold CV was performed on each of the five 3D CNN networks, the two 3D ResNet v1, the three 3D ResNet v2, and the two 3D DenseNet. Then, the network that achieved the highest average CV accuracy per model was retrained on the training set and evaluated on the test set. 

\subsubsection{Model architectures}
In this Section, all the model architectures were selected to fit within the GPU memory. The architectures of each of the five CNN models that were created using the monitored grid search are shown in Table \ref{models_reg_5}. These models architecture specifics were identical to those discussed in Section \ref{modelsArchitectures}. The only difference was the use of a ReLU activation function in all the networks' output layer. With respect to the 3D CNN models developed from scratch, Model 5 was the best performing model, since it achieved the best average CV MAE. Next, three 3D ResNet v1 networks were created, such that each network consists of one, two, and three residual blocks, respectively, and their depths are equal to $8$, $14$, and $20$ layers, respectively \citep{thesis}. The best performing model in the 5-fold CV for 3D RestNet v1 models was an architecture with a depth equal to $20$ layers.
Similarly, the next models investigated were two 3D ResNet v2 models, such that each network consists of one and two residual blocks, respectively, and their depths are equal to $11$ and $20$ layers,
respectively \citep{thesis}. Here, the best performing model in the 5-fold CV was an architecture with a depth equal to $11$ layers.
Finally, two 3D DenseNet networks were created, each having a 4-layer dense block and a 5-layer dense block with depth values equal to $23$ and $29$ layers, respectively. Per each model, each dense layer was preceded by a bottleneck layer \citep{thesis}, and each dense or bottleneck layer was followed by a dropout layer with a dropout rate equal to $0.2$ \citep{thesis}. The best performing 3D DenseNet model in the 5-fold CV was one with a depth equal to $23$ layers.

\begin{table}[h]
\caption[Architectures of the five 3D CNN models generated by the monitored grid search for the estimation of the total number of spikelets.]{Overall architectures of the five 3D CNN models generated by the monitored grid search for the estimation of the total number of spikelets.  \# of convolutional neuron is the number of neurons per convolutional layer, and \# of fully-connected neurons is the number of neurons per fully-connected layer.}

\label{models_reg_5}
\centering
\begin{tabular}{lcc}
\noalign{\smallskip}\noalign{\smallskip} 
\textbf{Model} & \textbf{\# of convolutional neurons} & \textbf{\# of fully-connected neurons} \\\hline
\noalign{\smallskip}
1 & 32,16 & 128,64,8 \\
2 & 32,32,8 & 128,16 \\
3 & 32,16,16,16 & 64,8 \\
4 & 32,16,8,8,8 & 32,32,16 \\
5 & 32,32,32,32 & 128 \\\hline
\end{tabular}
\end{table}

\subsubsection{Model Training}
\label{models_training_reg_sec}
Once 5-fold CV was completed, the model architectures that achieved the best average CV MAE were trained on the full training set and tested on the test set. Table \ref{models_reg_train} shows the training parameters (depth, optimizer, regularizer, epochs, and batch size) corresponding to the best performing models. Moreover, the optimizer, regularizer, epochs, and batch size also represent the training parameters for all the models. Starting with the 3D CNN, a batch size equal to $24$ was used. Adam optimization algorithm was used to update the network parameters with a learning rate equal to $1e^{-3}$, and MSE was used as the loss function. The 3D CNN was trained for $200$ epochs.

To train both 3D ResNet v1 and v2 models, an Adam optimizer was employed with a learning rate equal to $1e^{-3}$. Both models employed L2 regularizer with a regularization factor equal to $1e^{-4}$, and MSE as the loss function. 3D ResNet v1 and 3D ResNet v2 used a batch size equal to $12$ and $6$, respectively. Both models were trained for $200$ epochs.

Finally, to train 3D DenseNet, an Adam optimizer was employed with a learning rate equal to $1e^{-3}$ and a dropout regularizer was used. The model was trained for $200$ epochs with a batch size equal to $4$. An NVIDIA
Tesla P100 GPU server with 12GB of GPU memory was used to conduct all the experiments.
 
\begin{table}[h!]
\caption[The training parameters of the best performing architecture per model.]{The training parameters of the best performing architecture per model in the estimation of the total number of spikelets.}
\label{models_reg_train}
\centering
\begin{tabular}{lccccc}
\noalign{\smallskip}\noalign{\smallskip} 
\textbf{Model} & \textbf{Depth} & \textbf{Optimizer} & \textbf{Regularizer} & \textbf{Epochs} & \textbf{Batch size} \\\hline
\noalign{\smallskip}
3D CNN & 10 &  Adam & None & 200 & 24  \\
ResNet v1 & 20 &  Adam  & l2 & 200 & 12 \\
DenseNet & 23 &  Adam  & Dropout & 200 & 4  \\
ResNet v2 & 11 & Adam  & l2 & 200 & 6 \\\hline
\end{tabular}
\end{table}

\subsection{Model development for the estimation of the total number of infected spikelets}

\subsubsection{Monitored grid search}
\label{monitored_gs_infected}
The 3D CNN models created to estimate of the total number of infected spikelets on a wheat head were generated from scratch. The monitored grid search discussed in Sections \ref{monitored_grid_search} was followed. $100$ different networks were produced overall from five batches of $20$ networks. Only the architectures of the top three performing models will be discussed due to the large number of tested networks.

\subsubsection{Dataset characteristics, resizing, and labels}
\label{data_infected}
The data from Dataset II was used to train each of the models for the estimation of the total number of infected spikelets. Each of the PCs in the dataset was converted into a 3D image using a resolution factor $R=1.5$. Then, each 3D image was resized to ($227\times 70\times 111$) vx, which corresponds to its width, height, and depth. The labels of the data samples were integers between $2$ and $15$.

\subsubsection{5-fold cross validation}
\label{5fcv_infected}
Prior to training, the samples were divided into $80\%$ training set and $20\%$ test set. Next, the training samples were further divided into $5$ balanced folds to perform 5-fold CV, such that each fold consists of $20\%$ of the training data. We ensured a balanced and equal distribution of class labels between the train and the test sets and between the splits of the 5-fold CV. Next, a 5-fold CV was performed on each model following the same process as discussed in Section \ref{5fcv}.

\subsubsection{Model architectures}
For the estimation of the overall number of infected spikelets, a hundred 3D CNN models were trained in total. As a result of the vast number of tested models, only the architectures of the top three performers will be discussed, and which are depicted in Table \ref{models_infected}. The number of convolutional neurons refers to the number of neurons per 3D convolutional layer and the number of fully-connected neurons refers to the number of neurons per fully-connected layer. As mentioned in Section \ref{modelsArchitectures}, every 3D convolutional layer is followed by a 3D max pooling layer except for the last 3D convolutional layer. By default, the last fully-connected layer has $1$ neuron. As shown by Table \ref{models_infected}, Models 1 and 2 have identical 3D CNN architectures, with the optimizer being the only distinction, such that Model 1 has an Adam optimizer whereas Model 2 has a RMSprop optimizer. With respect to the 5-fold CV, Model 1 achieved the best average MAE amongst the top three best-performing models. Its architecture consists of four blocks, each consisting of a 3D convolutional layer and a 3D max pooling layer, where the kernel size of each convolutional layer is equal to $(3\times 3\times 3)$. Following the convolutional layers is a flattening layer, followed by three densely connected layers where the number of neurons per dense layer is equal to $32$, $8$, and $1$, respectively. The activation function in all the layers is the ReLU function.

\begin{table}[h]
\caption[Overall architectures of the top three 3D CNN models generated by the monitored grid search for the estimation of the number of infected spikelets.]{Overall architectures of the top three 3D CNN models generated by the monitored grid search for the estimation of the number of infected spikelets. \# of convolutional neuron is the number of neurons per convolutional layer, and \# of fully-connected neurons is the number of neurons per fully-connected layer.}
\label{models_infected}
\centering
\begin{tabular}{lccc}
\noalign{\smallskip}\noalign{\smallskip} 
\textbf{Model} & \textbf{\# of convolutional neurons} & \textbf{\# of fully-connected neurons} & \textbf{Optimizer} \\\hline
\noalign{\smallskip}
1 & 32,32,32,16 & 32,8 & Adam\\
2 & 32,32,32,16 & 32,8 & RMSprop\\
3 & 64,32,32,32 & 32 & Adam\\
\hline
\end{tabular}
\end{table}

\subsubsection{Model training}
To train each of the hundred 3D CNN models, a batch size equal to $4$ was used because there was not enough memory on the GPU to store a bigger batch. The RMSProp and the Adam optimization algorithms were either used to update the network’s parameters, with a learning rate within {$1e^{-4}, 0.5e^{-3},1e^{-3}$}, and MSE was used as a loss function. Each model was trained for $100$ epochs. A learning rate of $0.5e^{-3}$ was used for training Model 1.

\subsection{Model development for \emph{Fusarium} head blight severity assessment}

In this Section, FHB severity assessment was achieved by developing 3D CNN models that estimates directly the severity percentage. 
  

\subsubsection{Monitored grid search}
The 3D CNN models developed to estimate the FHB severity were built from scratch using the same methodology as described in Section \ref{monitored_gs_infected}. 

\subsubsection{Dataset characteristics, resizing, and labels}
The data used in this application is the same data that was used in the estimation of the total number of infected spikelets (see Section \ref{data_infected}). The labels were real values ranging from $92.3\%$ to $11.1\%$.

\subsubsection{5-fold cross validation}
The 5-fold CV process for the estimation of the FHB severity was the same as the process described in Section \ref{5fcv_infected}.

\subsubsection{Model architectures}
A total of a hundred 3D CNN models were trained for the estimation of the FHB severity of wheat infected with the FHB disease. Due to the large number of tested models, only the architectures of the top three performers, which are shown in Table \ref{models_SI}, will be discussed. The number of convolutional neurons refers to the number of neurons per 3D convolutional layer and the number of fully-connected neurons refers to the number of neurons per fully-connected layer. As mentioned in Section \ref{modelsArchitectures}, every 3D convolutional layer was followed by a 3D max pooling layer except for the last 3D convolutional layer. By default, the last fully-connected layer has 1 neuron. With respect to the results of the 5-fold CV, Model 1 achieved the best average MAE amongst the top three best-performing models. Its architecture consisted of four blocks, each consisting of a 3D convolutional layer and a 3D max pooling layer, where the number of neurons in each convolutional layer was $32$ except for the last layer, where the number of neurons was equal to $16$, and the kernel size in all the convolutional layers was equal to $(3\times 3\times 3)$. Following the convolutional layers was a flattening layer, followed by four densely connected layers where the number of neurons per dense layer was equal to $64$, $32$, $8$, and $1$, respectively. The activation function in all the layers was the ReLU function except for the output layer, where it used the sigmoid activation.

\begin{table}[h]
\caption[Overall architectures of the top three 3D CNN models generated by the monitored grid search for the estimation of the FHB severity of infected  wheat.]{Overall architectures of the top three 3D CNN models generated by the monitored grid search for the estimation of the FHB severity of infected wheat. \# of convolutional neuron is the number of neurons per convolutional layer, and \# of fully-connected neurons is the number of neurons per fully-connected layer.}
\label{models_SI}
\centering
\begin{tabular}{lcc}
\noalign{\smallskip}\noalign{\smallskip} 
\textbf{Model} & \textbf{\# of convolutional neurons} & \textbf{\# of fully-connected neurons}\\\hline
\noalign{\smallskip}
1 & 32,32,32,16 & 64,32,8 \\
2 & 32,32,32,32 & 64,32,8 \\
3 & 32,32,32,32 & 32 \\
\hline
\end{tabular}
\end{table}

\subsubsection{Model training}
To train each of the hundred 3D CNN models, a batch size equal to $4$ was used. The RMSProp optimization algorithms was used to update the network’s parameters, with a learning rate equal to $ 0.5e^{-3}$ and MSE was used as a loss function. Each model was trained for $100$ epochs. 

\subsection{FHB Disease assessment and statistical analysis}
FHB severity data collected by visual observations at $14$ DPI for nineteen ($19$) \emph{F. graminearum} isolates was analyzed using SAS Studio software version 3.8 (SAS Institute Inc., Cary, NC). A generalized linear mixed model with a beta distribution function was fitted to the data using PROC GLIMMIX with the LOGIT link function and BETA distribution (SAS, 2014). The isolates were treated as a fixed factor and replicated as a random factor. When a factor effect was significant, as indicated by a significant F test (p $\leq 0.05$), differences between the respective means were determined using the Least Significant Difference (LSD) test (p $\leq 0.05$). To determine the relationship between the results obtained by the 3D CNN model for severity estimation and the severity results collected by visual observation, randomly selected data collected at random DPI days, ranging from $4$ to $18$, from wheat plants inoculated with $19$ different \emph{F. graminearum} isolates and non-inoculated plants were used in a regression analysis using SAS Studio software.

\section{Results}

\subsection{Point cloud to 3D image conversion}
\label{Conversion_results}



\begin{figure}[h]
\begin{center}
\includegraphics[width=180mm]{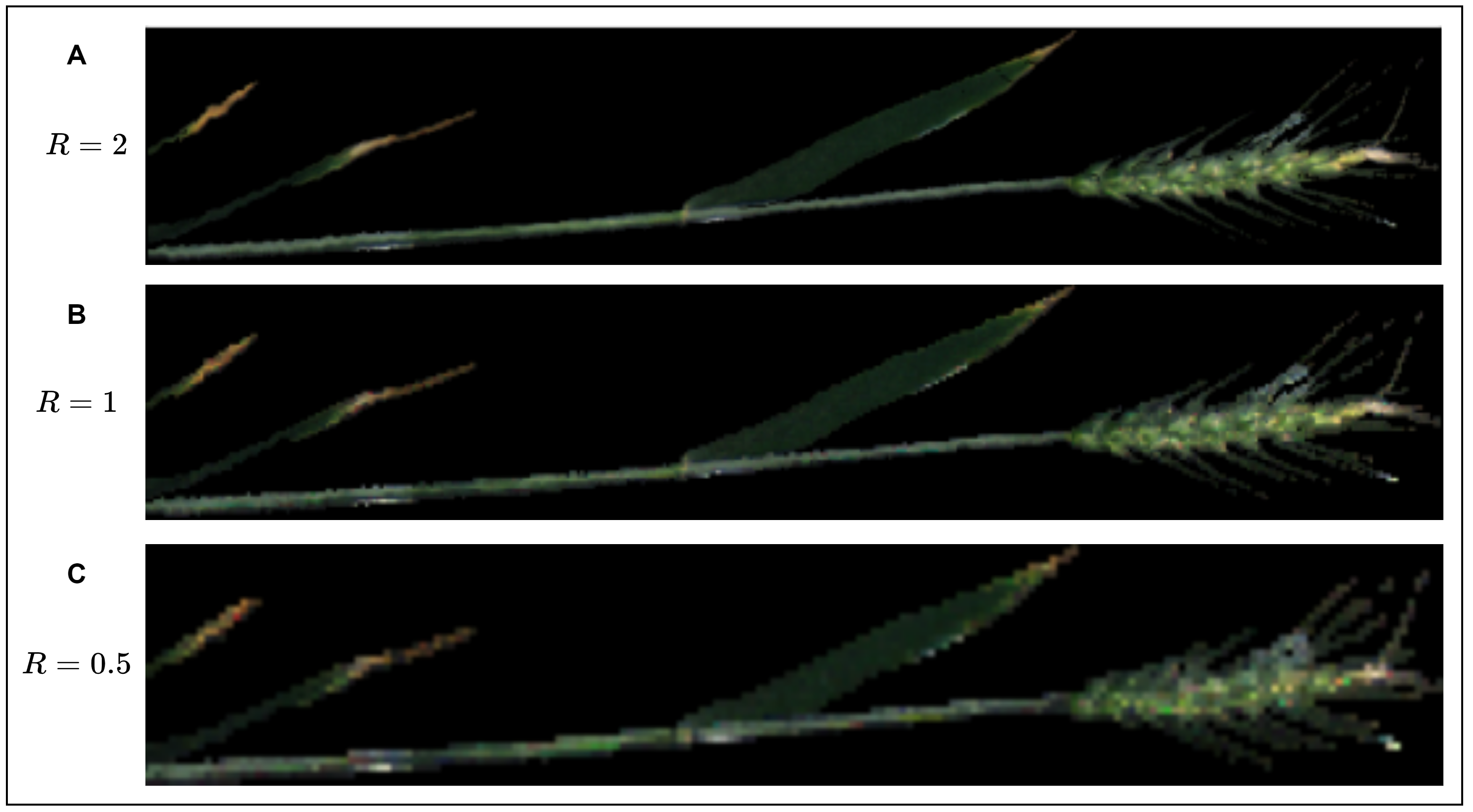}
\caption{2D projections of a 3D image converted with different resolution factors R using the CUDA
kernel.}
    \label{resolution}
\end{center}
\end{figure}

To visualize a few 3D images produced by the CUDA conversion model using various resolution factors, 2D projections were performed on 3D images representing a wheat plant. Figure \ref{resolution} shows three 2D projections of a wheat plant obtained by converting their original PLY files with different resolution factors ($R$). Figures \ref{resolution}(A), \ref{resolution}(B), and \ref{resolution}(C) depict images obtained from $R$ values equal to $2$, $1$, and $0.5$, respectively. The resolution of the images differ depending on the value of $R$, such that in Figure \ref{resolution}(C) where $R=0.5$, the image has a low-resolution due to the diminishing of the real dimensions by half during the conversion, whereas in Figure \ref{resolution}(B) where $R=1$, a higher resolution with more details and sharp edges can be observed due to the conservation of the real dimensions during the conversion. Figure \ref{resolution}(A), where $R=2$, shows a slightly better quality than Figure \ref{resolution}(B), such that its contours and details are more distinct.

\subsection{Detection of Fusarium head blight}
\label{FHBresults}
Table \ref{results} shows the detection performance metrics of the top three 3D CNN models over the three versions of datasets. Models 8, 10, and 11 achieved the highest average CV values amongst the batch of $20$ models on the 3DWP\_RGB dataset (characterized by RGB 3D images) by achieving $88.42\%$, $87.36\%$, and $86.84\%$ average CV accuracy, respectively. These three models were retrained and evaluated over the test set, and achieved $100\%$, $91.3\%$, and $91.3\%$ test accuracy, respectively. Models 11, 8, and 9 are the top three models amongst the $20$ models that attained the highest average CV accuracies on the 3DWP\_RGB\_NIR dataset (characterized by RGB${+}$NIR 3D images) and achieved $87.36\%$, $86.84\%$, and $85.78\%$, respectively. These three models were retrained and tested on the dataset’s test set, and despite Model 11 having the highest mean CV accuracy, it did not beat Model 8 in test accuracy. In fact, Model 8 achieved $95.65\%$ test accuracy, followed by Models 11 and 9 that achieved $91.3\%$ and $86.95\%$ accuracy, respectively. Finally, Models 3, 5, and 9 achieved the highest average CV accuracies on the 3DWP\_NIR dataset (characterized by NIR 3D images) by attaining $84.21\%$, $83.68\%$, and also $83.68\%$ average CV accuracy, respectively. Despite the fact that Model 9 achieved the lowest mean CV accuracy amongst the top three models, it obtained the highest test accuracy of $86.95\%$, followed by Models 5 and 3 that achieved $82.6\%$ and $78.26\%$, respectively. 

\subsection{Estimation of the total number of healthy and infected spikelets}

Table \ref{results} shows the results corresponding to the best performing models (3D CNN, 3D ResNet v1, 3D ResNet v2, and 3D DenseNet) in the regression problem on Dataset I of wheat heads. The table shows the performance metrics of the models, which are the average CV MAE, the test MAE, and the average prediction time per sample in milliseconds. Both the 3D CNN and ResNet v2 achieved the best test MAE of $1.13$. However, the 3D CNN outperformed the 3D ResNet v2 in the prediction time per sample with $14$ ms versus $112$ ms for the 3D RestNet v2. Moreover, even though 3D ResNet v1 obtained the best average CV MAE of $0.91$, it failed to obtain it on the test set with a MAE of $1.23$. However, 3D ResNet v1 produced the second-best prediction time of $62$ ms per sample. Although 3D DenseNet was ranked last in terms of average CV MAE in the group of models by obtaining a $1.28$ average CV MAE, it succeeded in achieving a $1.19$ MAE on the test set, which is ranked third. 3D DenseNet also achieved an average prediction time per sample of $140$ ms. 

\subsection{Estimation of the total number of infected spikelets}
Table \ref{results} shows the results corresponding to the best-performing 3D CNN models in the regression application, corresponding to the estimation of the total number of infected spikelets on the Dataset II. Amongst the hundred models that were tested, only three achieved the lowest MAE, which are depicted as Models 1, 2 and 3 in the Table. Model 1 achieved the best result of $1.56$ MAE, meaning that the predicted total number of infected spikelets in a wheat head is, on average, equal to the true label with a tolerance of $1.56$. Models 2 and 3 achieved the second- and third-lowest MAEs among the hundred models, which are equal to $1.57$ and $1.63$, respectively.

\subsection{Fusarium head blight severity assessment}
The results for the top-performing 3D CNN models in the regression application, which correspond to the estimation of FHB severity on Dataset II, are shown in Table \ref{results}. Only three of the $100$ models that were tested, identified in the Table as Models 1, 2, and 3, had the lowest MAE. Model 1 achieved the best result of $8.6$ MAE, meaning that the predicted FHB severity of a wheat plant is, on average, equal to the true label with a tolerance of $8.6\%$. The FHB severity value varies from $0\%$ (\emph{i.e.} all the spikelets are healthy) to $100\%$ (\emph{i.e.} all the spikelets are infected). Models 2 and 3 achieved the second- and third-lowest MAEs among the hundred models, which are equal to $8.8$ and $9.0$, respectively.

\begin{table}[h]
\caption[Evaluation metrics of the top CNN models per dataset and by application.]{Evaluation metrics of the top 3D CNN models per dataset and by application.}
\label{results}
\centering
\resizebox{\textwidth}{!}{
\begin{tabular}{lllccccc}
\noalign{\smallskip}\noalign{\smallskip}
\textbf{Application} & \textbf{Dataset} & \textbf{Models} & \textbf{Avg CV acc} & \textbf{Test acc \%} & \textbf{Avg CV MAE} & \textbf{Test MAE} & \textbf{Inference time} \\
\noalign{\smallskip}\hline\noalign{\smallskip} 
FHB detection & 3DWP\_RGB & Model 8 &  88.42  &   100  &- & -& -\\
 &  & Model 10 &  87.36  &  91.30  &- & -& -\\
 & & Model 11 &  86.84  &  91.30  &- & -& -\\
 & 3DWP\_RGB\_NIR & Model 11 &  87.36  &   91.30  &- & -& -\\
 &  & Model 8 &  86.84  &  95.65  &- & -& -\\
 & & Model 9 &  85.78  &  86.95  &- & -& -\\
   & 3DWP\_NIR & Model 3 &  84.21  &   78.26  &- & -& -\\
 &  & Model 5 &  83.68  &  82.60  &- & -& -\\
 & & Model 9 &  83.68  &  86.95  &- & -& -\\
\noalign{\smallskip}
Total \# of spikelets & Dataset I (heads) & 3D CNN & - & - &  1.26  &  1.13  &  14 \\
estimation & & ResNet v1 & - & - &  0.91  &   1.23   &  62 \\
 & & DenseNet & - & - &  1.28  &   1.19   &  140  \\
 & & ResNet v2 & - & - &  1.05 &  1.13   & l 42 \\
\noalign{\smallskip}
 \# of infected spikelets & Dataset II & Model 1 & - & - & 2.06 & 1.56 & - \\
estimation &  & Model 2 & - & - & 2.09 & 1.57 & - \\
&  & Model 3 & - & - & 3.05 & 1.63 & - \\
\noalign{\smallskip}
Severity estimation & Dataset II & Model 1 & - & - &12.4 & 8.6 &- \\
& & model 2 & - & - &12.6 & 8.8 &- \\
 & & model 3  & - & - &12.9& 9.0 &- \\
  \hline
  \multicolumn{8}{l}{\tiny{Average CV accuracy percentage (AVG CV acc \%), Test accuracy percentage (Test acc \%), and Average CV MAE (Avg CV MAE). }}\\
\end{tabular}}
\end{table}

\subsection{Visual FHB severity assessment vs automated assessment via 3D CNN}
\begin{figure}[h]
\begin{center}
\includegraphics[width=180mm]{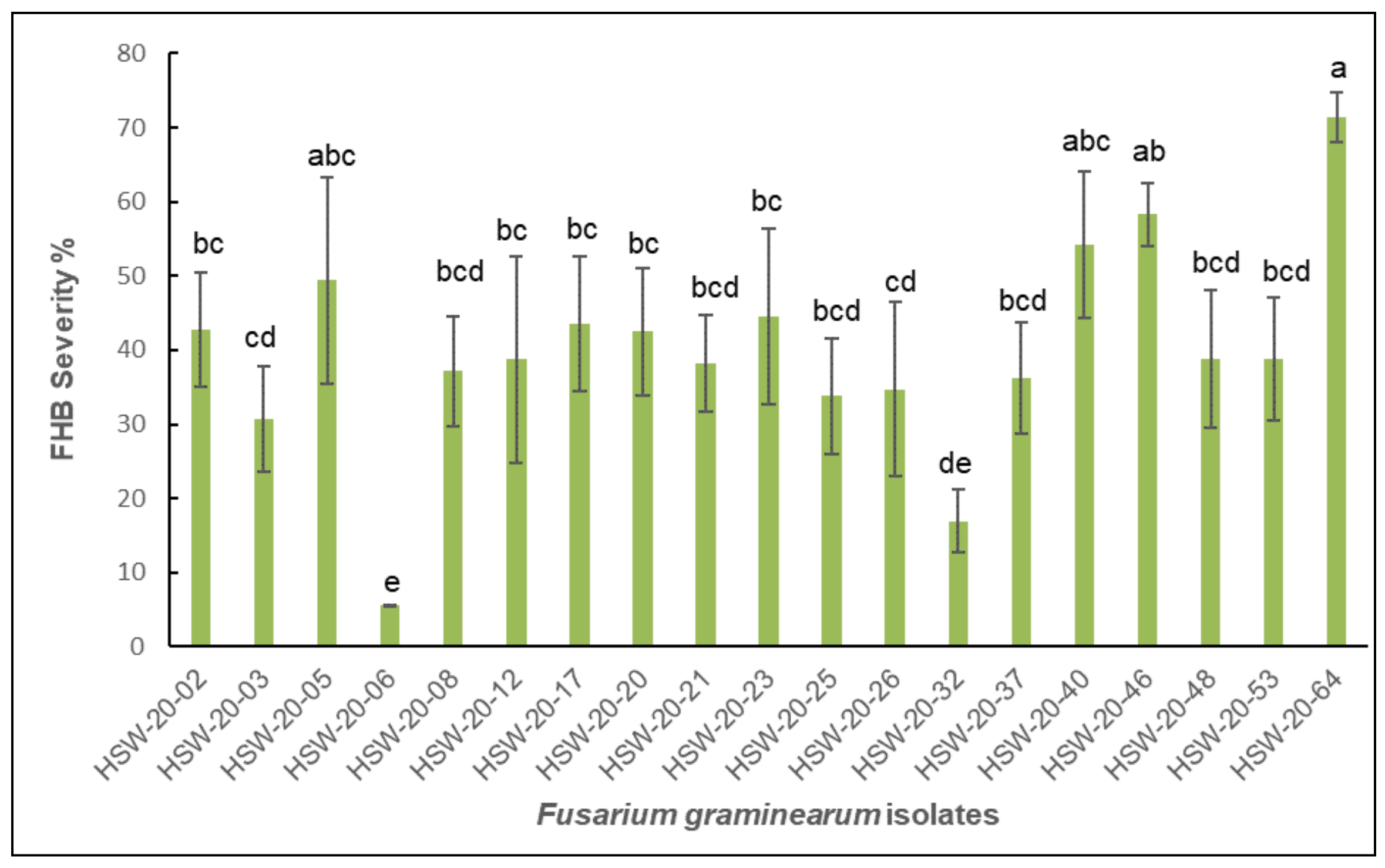}
\caption{Visual FHB disease assessment of the Canada Western Red Spring (CWRS) wheat cultivar 5602HR inoculated with \emph{F. graminearum} isolates at $14$ days post inoculation.  Means follow by a common letter are not statistically different at the $0.05\%$ level of significance according to Fisher's unprotected Least Significance Difference (LSD).}
    \label{isolates}
\end{center}
\end{figure}

\begin{figure}[h]
\begin{center}
\includegraphics[width=180mm]{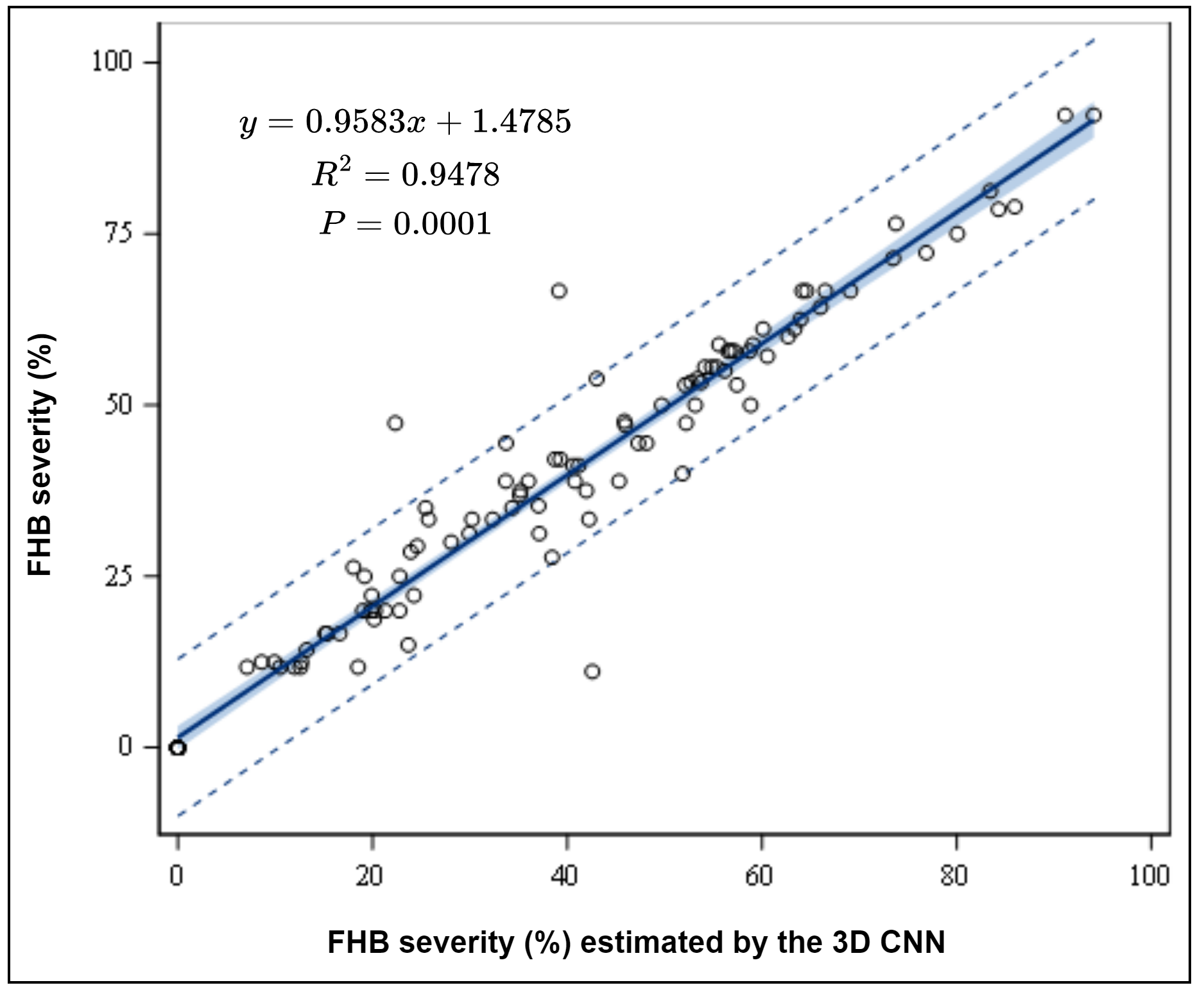}
\caption{Linear regression analysis of visual FHB disease assessment (Linear regression analysis of visual FHB disease assessment (FHB severity (\%)) and 3D convolutional neural networks Randomly selected wheat heads All the wheat heads, inoculated and non-inoculated, with $19$ \emph{F. graminearum} isolates at random DPI days, ranging from $4$ to $18$, were included in the analysis (n = $112$). The black solid line represents the fit line, the blue shaded area represents the $95\%$ confidence interval, and the dotted lines represent the prediction interval.}
    \label{CNNvsVISUAL}
\end{center}
\end{figure}

We performed visual assessments of the \emph{F. graminearum} infection at $7$, $14$, and $21$ dpi using a set of $19$ different \emph{F. graminearum} showing that all strains are pathogenic and a wide range of aggressiveness levels was observed. At $14$ dpi, there were significant differences among the \emph{F. graminearum} isolates inoculated into the wheat cultivar 5602HR (Figure \ref{isolates}). The FHB severity mean ranges from $5.6\%$ to $71.3\%$. Randomly selected wheat heads, both inoculated with the $19$ different \emph{F. graminearum} isolates and not inoculated, were used for linear regression. There was a significant relationship ($R^{2} = 0.94$, $P = 0.0001$) between the visual disease assessment and the data obtained with the 3D CNN model (Figure \ref{CNNvsVISUAL}).

\section{Discussion}
The present work shows the superiority of the RGB colour model over NIR, and RGB and NIR combined in the task of FHB detection in wheat plants. In fact, adding NIR information to RGB reduced the accuracy of the FHB detection from $100\%$ to $95.65\%$, which is a peculiar finding since, in general, adding more information to the data has the tendency to enrich it and give more information that should positively impact the performance of the CNN. However, in our case, NIR information is observed to have a negative influence on the learning performance of the classifiers from 3D images of wheat. Using data consisting of only NIR information resulted in the lowest accuracy of $86.95\%$ on the task of FHB detection in 3D images of wheat. Thus, we concluded that the RGB channels are the most efficient colour channels for the task of FHB detection, while NIR information does not enhance disease detection but rather reduces it. These findings may be explained by the fact that the symptoms of the FHB disease are clearly observed with the naked eye in wheat and that there are potentially no hidden symptoms that could be enhanced with the NIR spectrum. Therefore, NIR channel can provide misleading information to the models rather than any valuable details.

Moreover, our results obtained from the 3D CNN regression model in predicting the number of spikelets per wheat head are very promising. The true labels of Dataset I for wheat heads are integer values ranging between $7$ and $22$, yet the 3D CNN succeeded in efficiently
predicting these numbers with an error value of only $1.13$, meaning that, the difference between the predicted output and the real output is on average equal to $1.13$. Even though the MAE is not
negligible, it is still considered a very neat result that can be further improved, and it is still much better than the rough estimation that is manually performed by humans. 

Additionally, the top-performing 3D CNN model was reasonably successful in accurately predicting the right labels, with an MAE of $1.56$. The results are still within an acceptable error range despite the fact that the MAE is not negligible, and the automated tool can still be considered an efficient and a time-saving replacement for the manual and subjective calculation of the total number of infected spikelets.

Furthermore, the results obtained from the 3D CNN models trained for predicting the FHB severity are promising. The best-performing model achieved an $8.6$ MAE on the Dataset II. Despite the fact that MAE is not negligible, it can still be considered a tolerable margin of error, and the tool can be an efficient replacement for the manual estimations. 

Finally, the linear regression results of $R^{2} = 0.94$ and $P = 0.0001$, demonstrate the existence of a significant correlation between the FHB severity assessment using 3D CNN and the visual FHB severity. This confirms that automated assessment of the disease's severity is a successful means of determining FHB severity in infected wheat plants, even at very early stages of the infection. Moreover, this technology can be implemented and applied in different areas that focus on the management of FHB, such as plant breeding programs, precision crop protection, or the evaluation of fungicidal compounds.


\section*{Conflict of interest statement} 
The authors declare that the research was conducted in the absence of any commercial or financial relationships that could be construed as a potential conflict of interest.

\section*{Author contributions}
OH: Conceptualization, methodology, software, validation, investigation, data generation, writing - original draft, writing - review and editing, visualization.
CH: Conceptualization, validation, resources, data generation, writing - original draft, writing - review and editing, supervision, funding acquisition.
OM: Formal analysis, writing - review and editing, visualization.
CB: Resources, writing - review and editing, supervision, funding acquisition.
MH: Conceptualization, validation, formal analysis, resources, data generation, writing - review and editing, supervision.

\section*{FUNDING}
This work was funded by Mitacs Accelerate IT25876, Western Economic Diversification Canada Project No. 15453, and Agriculture and Agri-Food Canada.

\section*{Acknowledgements}
We thank Otto Gruenke and Debbie Miranda for their technical support in maintaining plants and preparing inoculations.

\section*{Data availability statement}
The original contributions presented in this study were produced using a public dataset created by the authors \citep{SP3/QJWBEM_2023}. The dataset is available at: \href{https://borealisdata.ca/dataset.xhtml?persistentId=doi:10.5683/SP3/QJWBEM}{https://borealisdata.ca/dataset.xhtml?persistentId=doi:10.5683/SP3/QJWBEM}.

\bibliographystyle{unsrtnat}
\bibliography{references}  






\end{document}